\documentclass[UTF8, conference]{IEEEtran}
\IEEEoverridecommandlockouts
\usepackage{CJKutf8}  % 替代 ctex，适用于 pdfLaTeX
\usepackage{amsmath,amssymb,amsfonts}
\usepackage{algorithmic}
\usepackage{graphicx}
\usepackage{textcomp}
\usepackage{xcolor}
\usepackage{pgfplots} %tikz 
\pgfplotsset{compat=1.17}
\usepackage{multirow}
\usepackage{booktabs} 
\usepackage[numbers]{natbib} % to use bib file
\usepackage{url}

\def\BibTeX{{\rm B\kern-.05em{\sc i\kern-.025em b}\kern-.08em
    T\kern-.1667em\lower.7ex\hbox{E}\kern-.125emX}}
\begin{document}

\title{Zero-Shot End-to-End Relation Extraction in Chinese: A Comparative Study of Gemini, LLaMA, and ChatGPT}

\author{
\IEEEauthorblockN{Shaoshuai Du$^*$}
\IEEEauthorblockA{\textit{University of Amsterdam} \\
\texttt{s.du@uva.nl}}
\and
\IEEEauthorblockN{Yiyi Tao}
\IEEEauthorblockA{\textit{Johns Hopkins University} \\
\texttt{ytao23@jhu.edu}}
\and
\IEEEauthorblockN{Yixian Shen}
\IEEEauthorblockA{\textit{University of Amsterdam} \\
\texttt{y.shen@uva.nl}}
\and
\IEEEauthorblockN{Hang Zhang}
\IEEEauthorblockA{\textit{University of California San Diego} \\
\texttt{haz006@ucsd.edu}}
\and
\IEEEauthorblockN{Yanxin Shen}
\IEEEauthorblockA{\textit{Simon Fraser University} \\
\texttt{yanxin\_shen@sfu.ca}}
\and
\IEEEauthorblockN{Xinyu Qiu}
\IEEEauthorblockA{\textit{Northeastern University} \\
\texttt{qiu.xiny@northeastern.edu}}
\and
\IEEEauthorblockN{Chuanqi Shi}
\IEEEauthorblockA{\textit{University of California San Diego} \\
\texttt{chs@ucsd.edu}}
}
\maketitle
\footnotetext[1]{*Corresponding author: Shaoshuai Du (s.du@uva.nl).}

\begin{abstract}
This study investigates the performance of various large language models (LLMs) on zero-shot end-to-end relation extraction (RE) in Chinese, a task that integrates entity recognition and relation extraction without requiring annotated data. While LLMs show promise for RE, most prior work focuses on English or assumes pre-annotated entities, leaving their effectiveness in Chinese RE largely unexplored.
To bridge this gap, we evaluate ChatGPT, Gemini, and LLaMA based on accuracy, efficiency, and adaptability. ChatGPT demonstrates the highest overall performance, balancing precision and recall, while Gemini achieves the fastest inference speed, making it suitable for real-time applications. LLaMA underperforms in both accuracy and latency, highlighting the need for further adaptation. 
Our findings provide insights into the strengths and limitations of LLMs for zero-shot Chinese RE, shedding light on trade-offs between accuracy and efficiency. This study serves as a foundation for future research aimed at improving LLM adaptability to complex linguistic tasks in Chinese NLP.
\end{abstract}


\section{Introduction}
With the rapid development of AI, numerous applications have emerged across various domains, including image processing, financial sentiment analysis, and natural language understanding~\cite{10055297/sdu,huang2024image2text2image/yixian,shen2024financialsentimentanalysisnews/yanxin,xu2024improving/shi,tao2024nevlp/luntao}. Among these, zero-shot end-to-end relation extraction (RE) has gained increasing attention as a promising approach for extracting structured knowledge from unstructured text without requiring task-specific training data.

Zero-shot RE aims to extract entities and their semantic relationships without annotated examples, relying solely on a model’s pre-trained knowledge and inference capabilities. Unlike traditional pipeline approaches that separate entity recognition and relation classification, end-to-end RE integrates these two processes into a single framework, reducing error propagation and improving contextual understanding. However, zero-shot end-to-end RE remains highly challenging, particularly in Chinese, due to:
\begin{itemize}
    \item \textbf{Lack of explicit word boundaries:} Unlike English, where words are separated by spaces, Chinese requires implicit segmentation, making entity boundary detection more ambiguous.
    \item \textbf{Complex character composition:} Many entity names share overlapping characters, increasing segmentation difficulty.
    \item \textbf{Implicit relation expression:} Chinese often conveys relationships through contextual inference rather than explicit syntactic patterns, making relation extraction harder.
\end{itemize}

Despite these challenges, recent advancements in Large Language Models (LLMs), such as GPT-4 and Gemini, have demonstrated strong capabilities in zero-shot learning, offering a potential solution for RE without annotated training data. While existing studies have primarily focused on English datasets, the effectiveness of different LLMs on zero-shot end-to-end RE in Chinese remains underexplored.

To bridge this gap, this study systematically evaluates the zero-shot end-to-end RE performance of three prominent LLMs: ChatGPT, Gemini, and LLaMA. Our contributions are as follows:
\begin{itemize}
    \item We provide a comparative analysis of these models’ capabilities in handling zero-shot Chinese RE, highlighting their respective strengths and weaknesses.
    \item We evaluate inference latency alongside accuracy metrics, offering insights into their practical feasibility.
    \item We introduce a semantic matching-based evaluation method to mitigate inconsistencies in entity recognition and relation phrasing.
\end{itemize}

This study advances the understanding of LLMs in zero-shot RE for Chinese, addressing both the challenges of linguistic complexity and the trade-offs between accuracy and efficiency.

The remainder of this paper is organized as follows: Section~\ref{sec:models_related_work} introduces the models and related work. Section~\ref{sec:methodology} outlines the methodology, including the workflow and evaluation metrics. Section~\ref{sec:experiments} presents the experiments, results, and analysis. Finally, Section~\ref{sec:conclusion} summarizes the contributions and future directions.

\section{Models and Related Work}
\label{sec:models_related_work}

\subsection{Models}
\textbf{ChatGPT}\cite{openai_chatgpt}, developed by OpenAI, excels in multilingual understanding and generating human-like text. Its pre-training makes it a strong candidate for zero-shot RE, but it is sensitive to prompt design and may produce verbose outputs without clear instructions.
\textbf{Gemini}\cite{google_gemini} is a multilingual LLM with Chinese support. We evaluate it using its API with prompts tailored to its input requirements. While versatile, Gemini may struggle with complex entity relations without fine-tuning due to limitations in its training data and architecture.
\textbf{LLaMA}\cite{meta_llama} is efficient and adaptable, with a Chinese-adapted version supporting structured outputs like triples. However, it may face challenges with nuanced entity recognition in Chinese without task-specific training.

Here are the specific models we used, as shown in Table~\ref{tab:models}

\begin{table}[h!]
    \centering
    \caption{Models Used for Comparison.}
    \resizebox{\columnwidth}{!}{
    \begin{tabular}{c|l|l|l}
    \toprule
    & \textbf{OpenAI} & \textbf{Google} & \textbf{Meta}\\
    \midrule
    \multirow{4}{*}{\centering \textbf{Models}}  & gpt-4 & gemini-1.0-pro & llama3.1-70b \\
    & gpt-4-turbo & gemini-1.5-flash & llama3.1-405b\\
    & gpt-4o-mini & gemini-1.5-flash-8b & llama3.2-3b\\
    & gpt-4o & gemini-1.5-pro & llama3.2-90b-vision \\
    \bottomrule
    \end{tabular}
    }
    \label{tab:models}
\end{table}

% \subsection{Related Work}
% LLMs have shown significant potential in zero-shot information extraction and knowledge graph construction. Wei et al.~\cite{wei2023chatie/chatie} introduced ChatIE, utilizing ChatGPT for zero-shot information extraction via conversational prompts. Zhu et al.~\cite{DBLP:journals/www/ZhuWCQOYDCZ24/llms} highlighted LLMs' capabilities in knowledge graph reasoning and construction, emphasizing their semantic understanding. Wu et al.~\cite{wu2024zero/medical} demonstrated the use of GPT-3.5-turbo and GPT-4 in zero-shot construction of Chinese medical knowledge graphs, addressing domain-specific challenges. However, most existing studies focus on LLMs released by one specific company, leaving the performance of other LLMs in zero-shot end-to-end RE for Chinese unexplored.

\subsection{Related Work}

Relation extraction is a fundamental NLP task that identifies semantic relationships between entities. Traditional \textbf{pipeline-based} RE treats entity recognition and relation classification as separate steps~\cite{DBLP:conf/acl/ZhouSZZ05/re}, leading to error propagation. In contrast, \textbf{end-to-end} RE integrates both tasks into a unified model, reducing error accumulation~\cite{DBLP:conf/emnlp/CabotN21/endtoend2}.

To address the reliance on labeled data, \textbf{zero-shot RE} leverages pre-trained knowledge to infer relations without task-specific annotations~\cite{DBLP:conf/conll/LevySCZ17/zeroshot}. Recent studies show that prompt-based zero-shot learning enables LLMs to extract relations from context~\cite{DBLP:journals/corr/abs-2307-01128/zeroshot2}. Wei et al.~\cite{wei2023chatie/chatie} used ChatGPT for conversational information extraction, while Zhu et al.~\cite{DBLP:journals/www/ZhuWCQOYDCZ24/llms} demonstrated LLMs’ capabilities in knowledge graph construction. Wu et al.~\cite{wu2024zero/medical} applied zero-shot RE to medical knowledge extraction using GPT models.

However, prior work primarily focuses on English datasets with pre-annotated entities, simplifying the task. Moreover, most evaluations center on OpenAI models, leaving the performance of alternatives like Gemini and LLaMA unexamined. More critically, zero-shot end-to-end RE in Chinese remains underexplored, despite its unique linguistic complexities.

To fill this gap, we systematically evaluate the zero-shot end-to-end RE performance of ChatGPT, Gemini, and LLaMA on Chinese datasets. Our study compares accuracy and inference efficiency while introducing semantic matching to enhance evaluation robustness beyond traditional string-based metrics.

\section{Methodology}
\label{sec:methodology}
\subsection{Workflow}

\begin{figure}[h!]
    \centering
    \includegraphics[width=\linewidth, trim=120pt 0pt 0pt 0pt, clip]{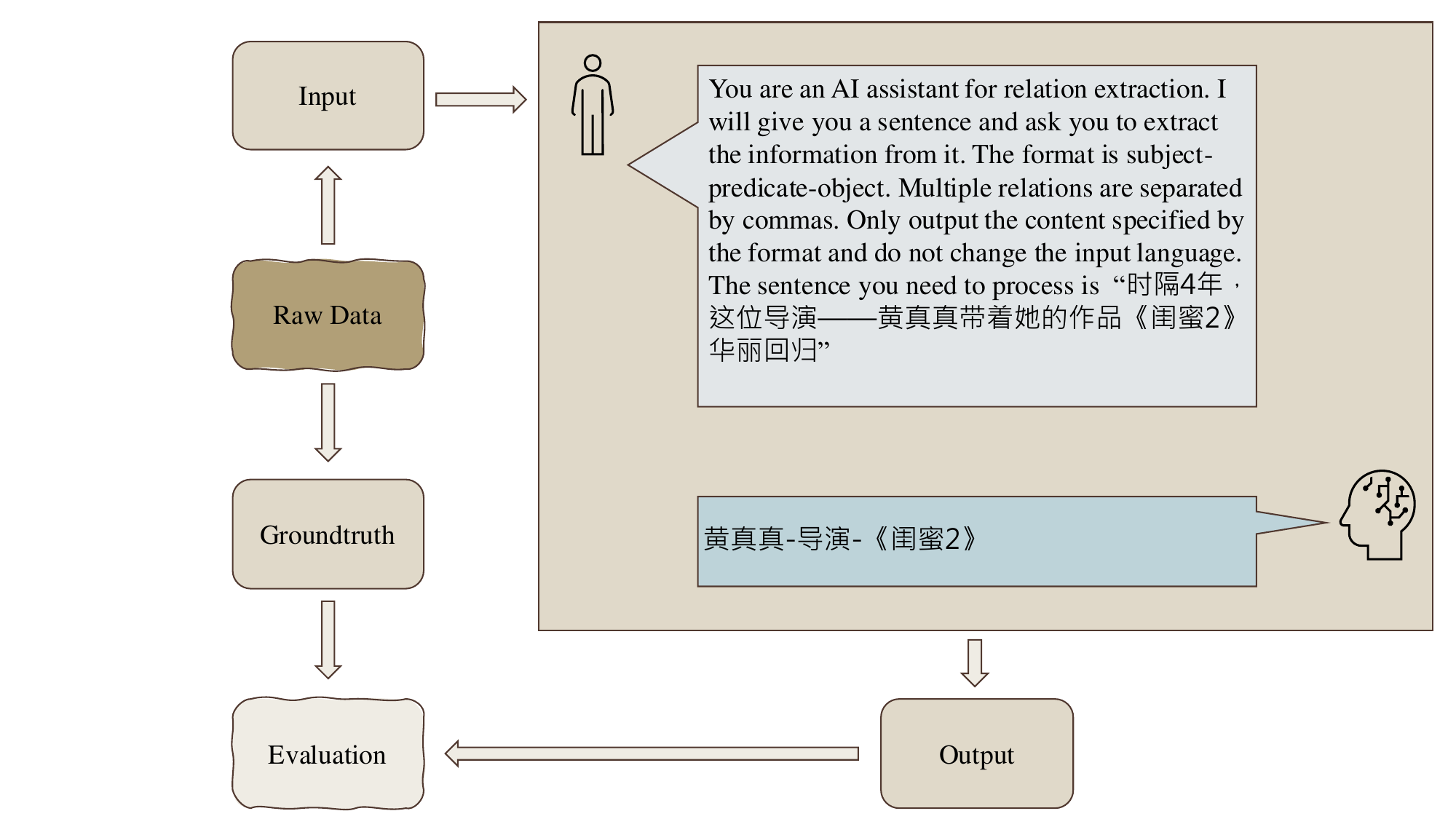}
    \caption{Workflow of Zero-Shot End-to-End Relation Extraction.}
    \label{fig:workflow}
\end{figure}

As illustrated in Figure \ref{fig:workflow}, the workflow begins with raw \textbf{data}, such as sentences from the DuIE 2.0 dataset, which are directly fed into the models' API without manual annotations. 
For example, given the prompt and the input sentence: \textit{\begin{CJK}{UTF8}{gbsn}“时隔4年，这位导演——黄真真带着她的作品《闺蜜2》华丽回归。”\end{CJK}} as shown in Figure~\ref{fig:workflow}, the models are prompted to extract entity-relation triples. In this case, an ideal output would be: \textit{\begin{CJK}{UTF8}{gbsn}“黄真真-导演-《闺蜜2》”\end{CJK}} %

Each model—Gemini, LLaMA, and ChatGPT—processes the input using its pre-trained knowledge, generating structured outputs in the form of triples. These outputs are then compared against the ground truth using joint evaluation metrics and semantic matching techniques, as detailed in Section~\ref{sec:metrics}.

\subsection{Joint Evaluation Metrics and Semantic Matching}
\label{sec:metrics}

Evaluating zero-shot end-to-end RE requires metrics that assess both entity recognition and relation extraction. Key metrics include:
\begin{itemize}
    \item \textbf{Joint Precision:} The proportion of correctly extracted entity-relation tuples out of all tuples extracted by the model.
    \item \textbf{Joint Recall:} The proportion of correctly extracted tuples out of all true tuples in the ground truth.
    \item \textbf{Joint F1 Score:} The harmonic mean of joint precision and recall, balancing accuracy and completeness.
\end{itemize}

To avoid the problem of semantic variability , evaluation methods incorporating semantic similarity measures have been introduced\cite{giunchiglia2003semantic}:
\begin{itemize}
    \item \textbf{Semantic Matching:} Techniques like word embeddings and contextualized models measure similarity between outputs and ground truth, accommodating variations in wording.
    \item \textbf{Relaxed Criteria:} Outputs are considered correct if they partially overlap with or provide acceptable alternatives to the ground truth.
\end{itemize}

% \subsection{Prompt Engineering for Zero-Shot End-to-End CN RE}
% An example prompt used for all models is as follows:

% "You are an AI assistant for relation extraction. I will give you a sentence and ask you to extract the information from it. The format is subject-predicate-object. Multiple relations are separated by commas. Only output the content specified by the format and do not change the input language. The sentence you need to process is: {Li Lei studies computer science at Peking University.}"

% This prompt instructs the model to: identify entities, determine the relationships between them and then output the results in a clear and structured format.

\section{Experiments}
\label{sec:experiments}

\subsection{Experimental Setup}

\textbf{API}: We utilize the latest APIs and libraries provided by each model’s developers. 

\textbf{Dataset} We use the DuIE 2.0 dataset\cite{DBLP:conf/nlpcc/LiHSJLJZLZ19/duie}, a widely recognized benchmark for Chinese information extraction tasks. It contains a diverse set of sentences annotated with entities and their relations, covering various domains and relation types.

\subsection{Results and Analysis}

\begin{table}[htbp]
\begin{center}
\caption{Performance Metrics of Models}
\resizebox{\columnwidth}{!}{
\begin{tabular}{l|c|c|c}
\toprule
\textbf{Model} & \textbf{Precision} & \textbf{Recall} & \textbf{F1} \\
\midrule
gpt-4 & \textbf{0.363} & 0.353 & 0.358 \\
gpt-4-turbo & 0.331 & 0.413 & \textbf{0.367} \\
gpt-4o-mini & 0.284 & \textbf{0.432} & 0.343 \\
gpt-4o & 0.308 & 0.223 & 0.258 \\
gemini-1.0-pro & 0.295 & 0.153 & 0.201 \\
gemini-1.5-flash & 0.231 & 0.424 & 0.293 \\
gemini-1.5-flash-8b & 0.330 & 0.206 & 0.254 \\
gemini-1.5-pro & 0.354 & 0.275 & 0.309 \\
llama3.1-70b & 0.249 & 0.134 & 0.174 \\
llama3.1-405b & 0.176 & 0.088 & 0.117 \\
llama3.2-3b & 0.215 & 0.108 & 0.143 \\
llama3.2-90b-vision & 0.268 & 0.133 & 0.178 \\
\bottomrule
\end{tabular}
}
\end{center}
\label{tab:performance_metrics}
\end{table}

%accuracy
\textbf{Precision, Recall and F1 Score}
In Table~\ref{tab:performance_metrics}, we present the precision, recall, and F1 scores of various models evaluated in our experiments. The results highlight that OpenAI's models, particularly \texttt{gpt-4-turbo}, demonstrate superior overall performance, achieving the highest F1 score of 0.367, which reflects a well-balanced trade-off between precision and recall. \texttt{gpt-4} achieves the highest precision at 0.363, indicating its strength in generating accurate predictions, with a slightly lower recall of 0.353. On the other hand, \texttt{gpt-4o-mini} achieves the highest recall of 0.432, showing strong coverage but at the expense of lower precision of 0.284.

The Gemini models exhibit mixed performance. For instance, \texttt{gemini-1.5-flash} shows a relatively high recall of 0.424, though its precision and F1 score remain moderate at 0.231 and 0.293, respectively. In contrast, Llama models generally perform poorly across all metrics. The \texttt{llama3.1-405b} model achieves the lowest F1 score of 0.117, indicating its limited effectiveness in the zero-shot CN RE task.

These results suggest that OpenAI's models, particularly \texttt{gpt-4-turbo}, are better suited for zero-shot relation extraction tasks compared to the Gemini and Llama series. However, the trade-offs between precision and recall in some models, such as \texttt{gpt-4o-mini}, indicate potential areas for optimization to improve the balance between prediction accuracy and coverage.

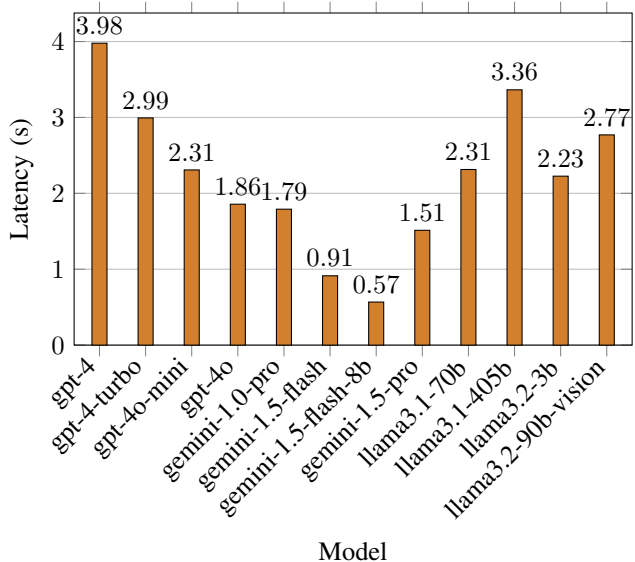
\begin{figure}[htbp]
\centering
\begin{tikzpicture}
    \begin{axis}[
        width= 9cm, 
        height=6cm, 
        ybar, 
        symbolic x coords={
            gpt-4, gpt-4-turbo, gpt-4o-mini, gpt-4o,
            gemini-1.0-pro, gemini-1.5-flash, gemini-1.5-flash-8b, gemini-1.5-pro,
            llama3.1-70b, llama3.1-405b, llama3.2-3b, llama3.2-90b-vision}, 
        xtick=data, 
        xlabel={Model}, 
        ylabel={Latency (s)}, 
        ymin=0, 
        bar width=0.2cm, 
        xticklabel style={rotate=45, anchor=east}, 
        nodes near coords, 
        nodes near coords align={vertical}, 
        enlarge x limits=0.05, 
        ymajorgrids, 
    ]
        \addplot[fill=brown!70!orange] coordinates {
            (gpt-4, 3.9782448)
            (gpt-4-turbo, 2.9932884)
            (gpt-4o-mini, 2.3082128)
            (gpt-4o, 1.8567588)
            (gemini-1.0-pro, 1.7909380)
            (gemini-1.5-flash, 0.9132564)
            (gemini-1.5-flash-8b, 0.5672590)
            (gemini-1.5-pro, 1.5128474)
            (llama3.1-70b, 2.3141026)
            (llama3.1-405b, 3.3638854)
            (llama3.2-3b, 2.2261884)
            (llama3.2-90b-vision, 2.7699876)
        };
    \end{axis}
\end{tikzpicture}
\caption{Latency Comparison of Models.}
\label{fig:latency_bar_chart}
\end{figure}
%latency
\textbf{Latency Comparison}
Beyond accuracy, inference latency is a crucial factor for practical deployment, as shown in Figure~\ref{fig:latency_bar_chart}. OpenAI's models generally exhibit higher latency, with \texttt{gpt-4} reaching 3.98 seconds, reflecting its computational complexity. In contrast, optimized versions such as \texttt{gpt-4o} and \texttt{gpt-4o-mini} achieve lower latencies of 1.86 and 2.31 seconds, respectively, making them more viable for real-time applications.

Gemini models offer the lowest latency, with \texttt{gemini-1.5-flash-8b} achieving 0.57 seconds, followed by \texttt{gemini-1.5-flash} at 0.91 seconds. This efficiency makes them well-suited for scenarios requiring fast response times, such as online reasoning tasks where real-time interaction is critical. However, their moderate accuracy suggests a trade-off between speed and extraction quality.

LLaMA models show varying latency, with \texttt{llama3.1-405b} reaching 3.36 seconds, comparable to \texttt{gpt-4}, while lighter versions such as \texttt{llama3.2-3b} and \texttt{llama3.2-90b-vision} demonstrate latencies of 2.23 and 2.77 seconds, respectively. 

\subsection{Discussion}
The findings reveal a trade-off between accuracy and inference speed in zero-shot end-to-end RE for Chinese. OpenAI's models, such as \texttt{gpt-4-turbo}, deliver strong extraction performance but have higher latency, limiting their suitability for real-time applications. In contrast, Gemini models, especially \texttt{gemini-1.5-flash-8b}, offer rapid inference, making them preferable for latency-sensitive applications like real-time decision-making. However, their moderate accuracy may impact performance in high-precision tasks. LLaMA models exhibit neither leading accuracy nor efficiency, suggesting a need for further adaptation. These insights highlight the importance of balancing accuracy and efficiency based on the specific requirements of real-world applications, whether in real-time processing or batch inference workflows.

\section{Conclusion}
\label{sec:conclusion}

This study evaluates ChatGPT, Gemini, and LLaMA for zero-shot end-to-end relation extraction in Chinese using semantic matching. ChatGPT achieves the best accuracy but suffers from higher latency, while Gemini offers the lowest latency with moderate performance. LLaMA underperforms across all metrics, indicating the need for further adaptation. These findings highlight the trade-offs between accuracy and efficiency in applying LLMs to Chinese RE.

Future work should explore finer-grained evaluation metrics, such as performance on specific relation categories and long-tail relations, to better assess model capabilities. Additionally, improving model adaptability and optimizing prompt design remain key directions for enhancing zero-shot Chinese RE.
\bibliographystyle{IEEEtran} 
\bibliography{references}

\end{document}